\documentclass{article}
\usepackage{ijcai22}

\usepackage{times}
\usepackage{soul}
\usepackage{url}
\usepackage[hidelinks]{hyperref}
\usepackage[utf8]{inputenc}
\usepackage[small]{caption}
\usepackage{graphicx}
\usepackage{amsmath}
\usepackage{amsthm}
\usepackage{booktabs}
\usepackage{algorithm}
\usepackage{algorithmic}
\usepackage{color}
\usepackage{xcolor}
\usepackage{adjustbox}

\usepackage{makecell}
\usepackage{multirow}

\title{Towards Federated Long-Tailed Learning}

\author{
Zihan Chen$^{1,2*}$ 
\and
Songshang Liu$^1$\thanks{Equal contributions. $^\dag$ Corresponding author.}\and
Hualiang Wang$^1$\and
Howard H. Yang$^1$ \and \\
Tony Q.S. Quek$^2$ \And
Zuozhu Liu$^{1\dag}$ 
\affiliations
$^1$ZJU-UIUC Institute, Zhejiang University, China\\
$^2$Singapore University of Technology and Design, Singapore
\emails
zihan\_chen@mymail.sutd.edu.sg,
songshang.17@intl.zju.edu.cn,
hualiang\_wang@zju.edu.cn,
haoyang@intl.zju.edu.cn,
tonyquek@sutd.edu.sg,
zuozhuliu@intl.zju.edu.cn.
}

\begin{document}

\maketitle

\begin{abstract}
   
   Data privacy and class imbalance are the norm rather than the exception in many machine learning tasks. Recent attempts have been launched to, on one side, address the problem of learning from pervasive private data, and on the other side, learn from long-tailed data. However, both assumptions might hold in practical applications, while an effective method to simultaneously alleviate both issues is yet under development. In this paper, we focus on learning with long-tailed (LT) data distributions under the context of the popular privacy-preserved federated learning (FL) framework. We characterize three scenarios with different local or global long-tailed data distributions in the FL framework, and highlight the corresponding challenges. The preliminary results under different scenarios reveal that substantial future work are of high necessity to better resolve the characterized federated long-tailed learning tasks.
   

\end{abstract}

\section{Introduction}
Federated learning (FL) has garnered increasing attentions from both academia and industries, as it provides an approach for multiple clients to collaboratively train a machine learning model without exposing their private data \cite{mcmahan2017communication,bonawitz2019towards}. 
This privacy-preserving feature has prevailed FL in a broad range of applications such as the healthcare, finance,  and recommendation systems \cite{andreux2020siloed,yang2020recommand}.
The data stem from different sources often exhibits a high level of heterogeneity, e.g., non-IID distribution and/or imbalance in the size, which impedes the FL performance \cite{li2020prox,wang2021addressing}. 
Although several methods have been proposed to circumvent this issue by tackling the drift and inconsistency between the server and clients \cite{wang2020nova,karimireddy2020scaffold}, 
the impacts from long-tailed data distribution, which is an extreme case of data heterogeneity and widely exists in the real world data (e.g., healthcare and user behaviors data \cite{kang2019decoupling,shang2022federated}), has yet been understood. 

\begin{figure}
    \includegraphics[width=1\columnwidth]{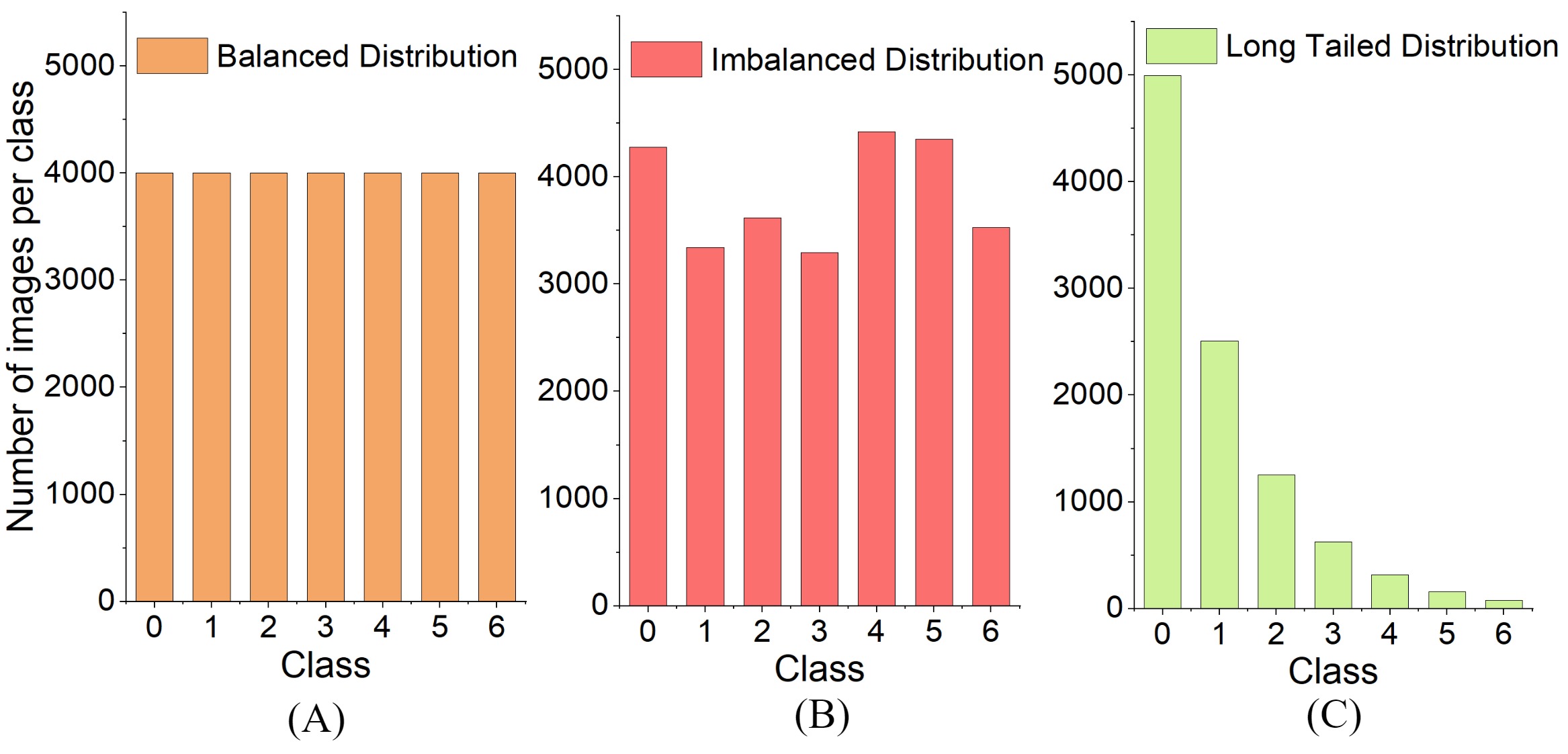}
    \caption{A comparison between the balanced, imbalanced, and long-tailed data distributions over a dataset with 7 classes. (A) is the balanced data distribution. (B) is the imbalanced data distribution. (C) is the long-tailed data distribution.}
    \label{fig:hist}
\end{figure}

Unlike data heterogeneity in the general sense, long-tailed distribution has a severely skewed shape in the distribution curve. 
To better illustrate this phenomenon, we provide a pictorial example in Figure~\ref{fig:hist}.
We differentiate the term long-tailed distribution from the category of imbalanced distribution in this figure as well as the rest of this paper to emphasize its unique role. 
Using Figure~\ref{fig:hist}, we can easily conclude that that in the presence of long-tailed data, training an unbiased classification model is generally challenging since the most of training data is concentrated in a few classes (i.e., the head classes) while the other classes (i.e., the tail classes) have very few samples. 
And it has been shown in  \cite{kang2019decoupling} that conventional deep learning models admit a significant performance degradation on real-world data that has a long-tailed distribution. 
In response, several schemes have been proposed to address such an extreme class imbalance issue. 
These methods are commonly known as the \textit{long-tailed learning}, established via the particular means of re-balancing \cite{zhang2021deep}, re-weighting \cite{lin2017focal}, and transfer learning techniques  \cite{yin2019feature}.  
Recently, decoupled representation and classification learning scheme \cite{kang2019decoupling} is investigated to effectively complement the conventional approaches (e.g., class-balanced sampling \cite{wang2020devil} and distribution-aware loss \cite{lin2017focal}). 

However, these existing solutions are primarily dedicated to the centralized learning (CL) and cannot be directly extended to the FL settings. 
Specifically, due to the distributed nature of the local data, it is much more difficult to train an unbiased model with the existence of long-tailed data in FL systems.
Additionally, the limited local dataset sizes of the local clients as well as the inherent data heterogeneity in FL also constrain the applicability of the approaches developed in the scenarios of CL \cite{yoon2020fedmix}.

We refer to the FL task with long-tailed data as the \textit{federated long-tailed learning.}
Note that long-tailed data distribution may exist in both the local and global level, leading to different challenges during the training procedure.
Particularly, the long-tailed  data distribution presents an obvious characteristic on the head and tail over different classes (See Figure \ref{fig:hist}~(C)). 
In FL systems, different clients could have different long-tailed properties and the overall (global) data distribution would also be balanced or imbalanced in different networks. 
The distribution of the real-world datasets is closely related to the user habits and geo-locations,
such as the image recognition datasets of the natural specifies (e.g., iNaturalist \cite{van2018inaturalist}) and the landmarks (e.g., Google Landmarks \cite{weyand2020landmarks}).
Such datasets would have a strongly geographical-dominated long-tailed distribution, and more importantly, images from different clients (in different locations) would present different distributional statistics.
It would be more challenging to train models with good generalization on different local long-tailed data distributions  than the single-distribution case.

Motivated by the aforementioned issues and the intrinsic properties of federated long-tail learning, 
the present paper gives a comprehensive analysis to the effect of long-tailed data on both the local and global level of FL, as well as the consequent challenges. In addition, numerical results in different settings are also provided to demonstrate the influence of long-tailed data distribution. Based on this, several future trends and open research opportunities are also discussed.

\section{Problem Formulation of Federated Long-Tailed Learning}

In this section, we will systematically characterize the Federated Long-Tailed (F-LT) learning problem, with the main difference lies at the distributions of the local data in each FL client and the aggregated global data distributions. The challenges under each setting are also discussed in detail. 

\subsection{Local and global data distribution}
Consider an FL system with $N$ clients and an $M$-class visual recognition dataset for classification problems, where $\mathcal{D}_k$ represents the local dataset for client $k$.
Let $n_k$ denote the size of the local dataset for client $k$ (i.e., $|\mathcal{D}_k|$), and $n_k^{(i)}$ denote the number of data samples of class $i$ in $\mathcal{D}_k$, i.e., $n_k = \sum_{i=1}^M n_k^{(i)}$.  

For a given client  $k$, we shall define the \textit{local data distribution} as 
\begin{equation}
    \mathbf{p}_k=[\frac{n_k^{(1)}}{n_k}, \dots, \frac{n_k^{(j)}}{n_k}, \dots, \frac{n_k^{(M)}}{n_k}],
\end{equation}
 where $\frac{n_k^{(j)}}{n_k}$ denotes the ratio of the $j$-th class over the corresponding local dataset size of client $k$. 

Note that in a typical FL system, the global server does not hold any data.  
To better capture the overall  data distribution from the system level, we define the \textit{global data distribution} as the distribution of the aggregated dataset from all clients in the system,  which is denoted by 
\begin{equation}
    \mathbf{p}_\text{G}=[\frac{\sum_{k=1}^N n_k^{(1)}}{|\mathcal{D}|},\frac{\sum_{k=1}^N n_k^{(2)}}{|\mathcal{D}|},\dots, \frac{\sum_{k=1}^N n_k^{(M)}}{|\mathcal{D}|}],
\end{equation}
where $|\mathcal{D}|=\sum_{k=1}^N n_k$ is the total number of samples in the FL system. 

Based on these two length-$M$ vectors $\mathbf{p}_k$ and $\mathbf{p}_\text{G}$, we can illustrate and analyze the distributional statistics of the long-tailed data from both the local and global perspectives. Specifically, the metric \textit{imbalance factor} (IF) \cite{zhou2020bbn,kang2019decoupling} could be used to measure the degree of long-tailed data distribution.
Given the local data distribution vector, the local imbalance factor for client $k$ is calculated by 
\begin{equation}
  \text{IF}_\text{L}^{(k)}=\frac{\max_j \{ n_k^{(j)}\} }{\min_s \{ n_k^{(s)}\} }.   
\end{equation}
Similarly, the  global imbalance  factor shall be denoted as
\begin{equation}
    \text{IF}_\text{G}=\frac{\max_j \{ \sum_{i=1}^N n_i^{(j)}\} }{\min_s \{\sum_{i=1}^N n_i^{(s)}\}}.
\end{equation}

\begin{figure*}
    \centering
    \includegraphics[width=2\columnwidth]{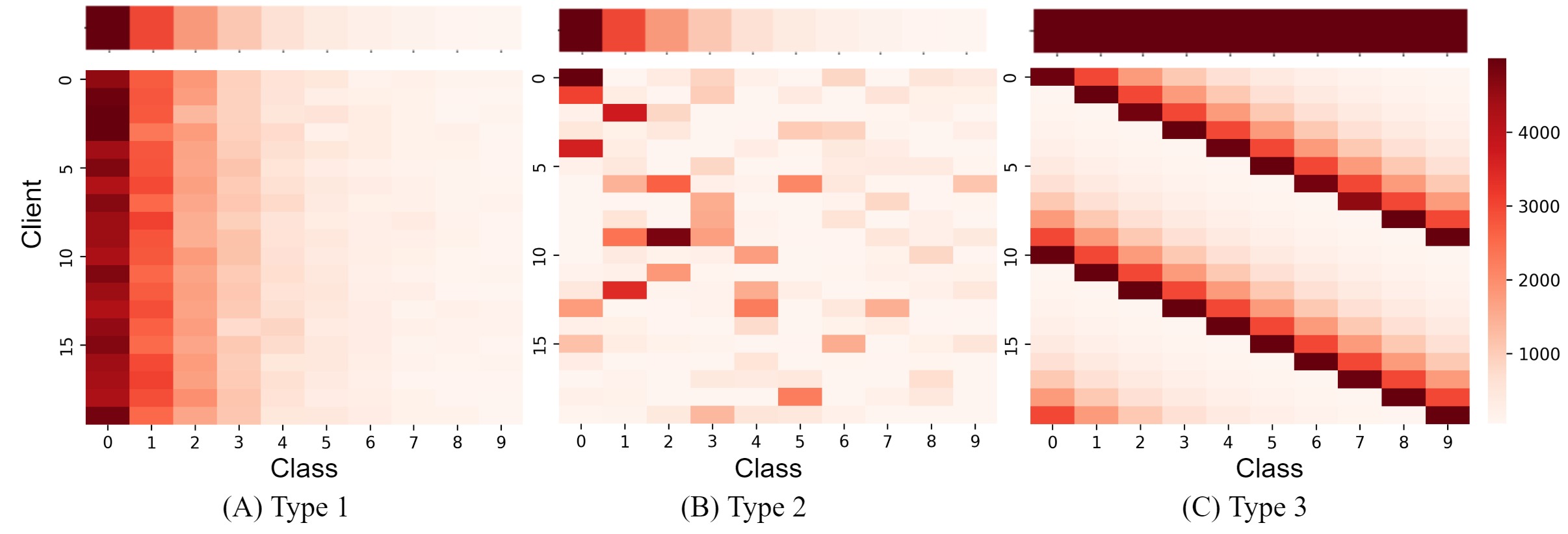}
    \caption{An example of the data distributions for the  summarized three types in a 20-clients FL system. The first row is the global data distribution in the corresponding type, with a colorbar in the right indicating the number of data samples in each class. Each sub-colorbox represents the number of data samples of each class across all clients.}
    \label{fig:heatmap}
\end{figure*}

\begin{table*}[t!]
\centering
\begin{tabular}{l|l|l|l}
\hline
\addlinespace[0.05cm]
Global data distribution &Local data distributions  &Objective of learning tasks  &Datasets    \\ \hline
\addlinespace[0.05cm]

\multirow{3}{*}{Long-tailed } &Identical long-tailed &  &Long-tailed datasets \\
&  distributions & Learn a good global model &  (e.g., CIFAR-10-LT) \\ \cline{2-4}
\addlinespace[0.05cm]
              &Long-tailed/ Imbalance/ &  & Long-tailed datasets \\
                & Balanced distibution  & Learn multiple good local models & (e.g., CIFAR-10-LT)  \\ \hline
\addlinespace[0.05cm]                 
\multirow{2}{*}{Non long-tailed } & Diversified long-tailed &  & Balanced datasets  \\ 
                  &distributions  & Learn multiple good local models  & (e.g., CIFAR-10) \\ \hline
\end{tabular}
\caption{A taxonomy of long-tailed data distribution in FL. The  objectives and potential datasets for the corresponding cases in federated long-tail learning are also provided.  }
\label{Tab:Summary}
\end{table*}

\subsection{Local and global long-tailed data distribution}
Note that either $ \text{IF}_\text{L}^{(k)}$ or $\text{IF}_\text{G}$ would be a large number in real-world datasets, which indicates that the long-tailed data distribution may exist in either the local side or global side. For example, the local medical image datasets in hospitals in a big city might follow long-tailed local distributions, while the aggregated city-level global dataset might be long-tailed or non long-tailed. 
Therefore,  considering the relations and differences between the local and global data distributions, we would categorize the federated long-tailed learning tasks into the following three types:
\begin{itemize}
    \item \textbf{Type 1: Both the local and  global data distribution follow the same long-tailed distribution.} In a homogeneous network, local data from all the clients follow the same distribution. In such a case, if the local data distribution has the long-tail characteristic, then the global data distribution would also be an identical long-tailed distribution. 
    \item \textbf{Type 2: Global data distribution is long-tailed, while local data distributions are diverse, and not necessarily long-tailed.} Local data of different clients in a heterogeneous network would be typically non-IID, where the pattern of the local data distribution would be rarely identical. Given a global long-tailed data distribution, the local data distributions of different clients could be long-tailed, imbalanced or balanced. 
    \item \textbf{Type 3: All or a subset of local clients have long-tailed data distributions, but the global data follows a non long-tailed distribution (e.g., balanced distribution over all classes).}  In the case that the global data distribution is non long-tailed, the pattern of the local long-tailed data distributions of different clients would be diverse (i.e., different clients are supposed to keep different head and tail classes.).
\end{itemize}

Incorporating the data heterogeneity (i.e., the non-IID and imbalanced dataset size), the overall three cases represent all possible scenarios of long-tailed data in a typical FL system.
As illustrated in Figure \ref{fig:heatmap},  we provide an example of the summarized three types for better visualization of the local and global distributions in federated long-tailed learning.

\subsection{Objective of learning tasks and potential approaches} 
\label{subsec:objective}
With the existence of long-tailed data distributions in FL systems, different cases would bring different challenges to the distributed learning process. We will discuss the characterized three types one by one respectively.

In the first type of long-tailed data distribution, local and global data distributions share the same statistical characteristics. 
A single well-trained global model has the potential to be well generalized over the local data from different clients in FL systems. 
As the long-tailed distributions of all clients are the same, one classifier trained for long-tailed data could be applicable for all clients.  Nevertheless, potential issues may arise due to the limited local dataset sizes.

In the remaining two types, a single distribution could not cover all possible distributions of the clients in the FL system.  
Conventional approaches for long-tail learning for a single  long-tailed distribution may fail to tackle such diversity issues.
We shall consider different learning objectives for different cases of local and global data distributions.
Specifically,  different local clients could have vastly diverse distributions (e.g., long-tailed and  non long-tailed), and the global and local data distributions would be different. 
Thus, it is necessary to train multiple models to address such discrepancies of data distributions. 

Recall that, in the context of the personalized federated learning (PFL) \cite{tan2022towardspfl}, personalized models for each client are trained, as one global model cannot be well generalized to diverse local clients. 
It would be natural to regard PFL as a key ingredient to tackle such diverse data distribution issues in these two scenarios. For example, a popular solution of PFL is to decouple the local model into base layers and personalization layers \cite{arivazhagan2019fedper}.
Recent works in the centralized long-tail learning demonstrate that decoupling the representation learning and classifier learning with a re-adjustment on classifier could effectively improve the performance \cite{kang2019decoupling,zhou2020bbn}. 
Such similar decoupling approaches on model parameters would intuitively make PFL approaches to complement the federated long-tail learning.

From a more general explanation, the key idea of the PFL is to find a good trade-off to balance the global shared knowledge and the local task-specific knowledge for personalized local training. Such a learning procedure could be applied to learn unbiased long-tail classifiers with a good generalizable representation. Moreover, multi-task learning (MTL) \cite{smith2017fmtl}, clustering \cite{ghosh2020clustering} and transfer learning approaches \cite{gao2019ppftl} could also have the potential to be applied to cross-device long-tail learning in FL, which shall be discussed later in detail (See Sec. \ref{Sec:Future}).

\renewcommand\arraystretch{1.3} 
\begin{table*}[t!]
\centering
\begin{adjustbox}{width=\textwidth,center}
\begin{tabular}{l|ccc|ccc|ccc|ccc}
\hline
 & \multicolumn{3}{c|}{Non-LT (IF$_\text{G}=1$)}  & \multicolumn{3}{c|}{IF$_\text{G}=10$} & \multicolumn{3}{c|}{IF$_\text{G}=50$} & \multicolumn{3}{c}{IF$_\text{G}=100$} \\ \hline
\multirow{2}{*}{Data Setting} & IID & \multicolumn{2}{c|}{Non-IID} & IID & \multicolumn{2}{c|}{Non-IID} & IID & \multicolumn{2}{c|}{Non-IID} & IID & \multicolumn{2}{c}{Non-IID} \\ \cline{2-13} 
 & - & $\alpha$=1 &$\alpha$=0.5 & - & $\alpha$=1 & $\alpha$=0.5 & - & $\alpha$=1 & $\alpha$=0.5 & - & $\alpha$=1 & $\alpha$=0.5 \\ \hline
FedAvg &0.9369  & 0.9316 & 0.9249 &0.8806  & 0.8761 & 0.8669 & 0.797 & 0.7863 & 0.7689 & 0.7393 & 0.7525 &0.7205  \\
FedProx&0.9382  & 0.9327 & 0.9275 & 0.8801 & 0.8785 & 0.8656 & 0.7943 & 0.7783 & 0.7719  & 0.7366 & 0.7499 & 0.7155 \\
CReFF& 0.945  & 0.9383 & 0.931 & 0.8914 & 0.8791 &  0.8736 &0.8059  &0.7953 &0.78  &0.7427  &0.7311 & 0.7118 \\
FedPer&0.9356  & 0.9296 & 0.9259 &0.8803  & 0.873 & 0.8696   &0.7633 &0.7503 &  0.7478  & 0.7376 & 0.7358 &0.7145  \\ \hline
\end{tabular}
\end{adjustbox}
\caption{Test accuracies of various FL methods on CIFAR-10-LT with different federated data partitions (i.e., Type 2). Results on balanced CIFAR-10 are also provided for reference and comparison.}
\label{Tab:CIFAR-LT}
\end{table*}

\begin{table}[t!]
\centering
\begin{tabular}{l|c|c|c}
\hline
Local Setting & IF$_\text{L}$ = 10 & IF$_\text{L}$ = 50 & IF$_\text{L}$ = 100 \\ \hline
FedAvg & 0.8896 & 0.859 & 0.8422 \\
FedProx & 0.8929 & 0.8586 & 0.8444 \\
CReFF& 0.8984& 0.8646 & 0.8485\\
FedPer& 0.8951 &0.8602  & 0.8438 \\ \hline
\end{tabular}
\caption{Test accuracies on CIFAR-10 with different local long-tailed distributions (i.e.,Type 3). }
\label{CIFAR}
\end{table}

\section{Benchmarking the Federated Long-Tailed Learning}
To the best of  our knowledge, the long-tailed learning in the context of FL has been rarely explored.  
In this section, we will give a summary on the datasets and the corresponding federated partition approaches. 
Recent works on long-tail learning in both centralized and federated scenarios will then be discussed.
At last, we would give a brief comparison on the two typical long-tailed data settings.

\subsection{Datasets and partition methods}
\noindent\textbf{Datasets} In a centralized paradigm for visual recognition tasks, there are mainly two types of dataset benchmarking for long-tailed study. The first type is the long-tailed version of image datasets modified with synthetic operation, such as exponential sampling (CIFAR10/100-LT \cite{cao2019learning}) and Pareto sampling( ImageNet-LT  \cite{openlongtailrecognition}, Places-LT \cite{openlongtailrecognition}). They are shaped/sampled from the existing balanced dataset and the degree of the long-tail could be controlled with an arbitrary imbalance factor IF$_\text{G}$.  Second type is the real-world large scale datasets with a highly imbalanced label distribution, like iNaturalist \cite{van2018inaturalist}  and Google Landmarks \cite{weyand2020landmarks}. More long-tailed datasets are used in some specific tasks, such as object detection Lvis \cite{gupta2019lvis}, multi-label classification VOC-MLT \cite{wu2020distribution} and COCO-MLT \cite{wu2020distribution}.



\noindent\textbf{Partition methods for long-tailed FL} 
To create different federated (distributed) datasets according to the different patterns of local and global data distribution, different datasets and sampling methods are required. 
Data distributions in Type 1 could be realized by IID sampling on long-tailed datasets. 
Similarly, Type 2 could be achieved by Dirichlet-distribution \cite{hsu2019measuring} based generation method on the long-tailed datasets. 
Specifically, the degree of the long-tail and the identicalness of local data distributions could be controlled by the global imbalance factor IF$_\text{G}$ and the concentration parameter $\alpha$ respectively. And Type 3 could be realized via the different long-tailed sampling (different head and tail pattern) on the balanced datasets. 

\subsection{Approaches}
\noindent\textbf{Centralized long-tail learning} 
In the centralized scenario, long-tailed learning seeks to address the class imbalance in training data. The most direct way is to rebalance the samples of different classes during the model training, such as ROS and RUS \cite{zhang2021deep}, Simple calibration \cite{wang2020devil} and dynamic curriculum learning \cite{wang2019dynamic}. The balancing ideology could also be implemented in re-weighting and remargining the loss function, such as Focal Loss \cite{lin2017focal}, LDAM Loss \cite{cao2019learning}. These class rebalancing methods could improve the tail performance at the expense of head performance.

To address the limitation of information shortage, some studies focus on improving the tail performance by introducing additional information, such as transfer learning, meta learning, and network architecture improvement. In transfer learning, there have been methods FTL \cite{yin2019feature} and LEAP \cite{liu2020deep} transferring the knowledge from head classes to boost the performance in tail classes. In \cite{shu2019meta}, meta-learning is empirically proved to be capable of adaptively learning an explicit weighting function directly from data, which guarantees robust deep learning in front of training data bias. Recently, some studies design and improve network architecture specific to long-tailed data. For example, different types of classifiers are proposed to address long-tailed problems, such as $\tau-$norm classifier  \cite{kang2019decoupling} and Causal classifier \cite{tang2020long}.


\noindent\textbf{Federated long-tail learning}
Yet, the only one related work on federated long-tail learning \cite{shang2022federated} utilized classifier re-training to re-adjust decision boundaries, where the discussion is limited within the global long-tailed distribution with local heterogeneity. 
Methods for other types of local and global data distribution remain to be further explored.

Nevertheless, in the presence of long-tailed data, the discrepancies among local and global data distributions of different clients in the FL system, could be possibly addressed by the techniques in the federated optimization algorithm, such as dynamic regularization \cite{acar2021feddyn}, diverse client scheduling \cite{cho2020poc} and adaptive aggregation. 
In addition, as we discussed previously in Sec. \ref{subsec:objective}, PFL could be applied in federated long-tailed learning to find a balance between the representation and the classification learning.
We shall give a detailed discussion on such explorations to boost the performance of federated long-tailed learning in Sec. \ref{Sec:Future}.

Based on the above discussion about the data distribution, datasets and learning  objectives, we summarize them into Table \ref{Tab:Summary}. 
Note that, the case, where both the local and global data distributions are non-long-tailed, is not listed in this table, as this case is not within the scope of this paper.

\subsection{Performance comparison}
To better illustrate the impacts of the long-tail data distribution, we shall provide some numerical results with different types of long-tailed data distribution in Tables \ref{Tab:CIFAR-LT} and \ref{CIFAR}. 
For all the experiments, we consider a FL with $40$ clients.
And the  non-IID data partition is implemented by Dirichlet distribution.
Apart from the basedline FedAvg \cite{mcmahan2017communication}, the other three FL algorithms are  FedProx \cite{li2020prox}, CReFF \cite{shang2022federated} and FedPer \cite{arivazhagan2019fedper}, which are representative approaches to tackle data heterogeneity, long-tailed data and  personalization in FL respectively.

Note that, the main purpose of this subsection is to analyze the performance of the different FL methods with diverse data settings to provide some  possible insights to the design of the federated long-tailed learning algorithm.

We choose two typical long-tailed data distributions in the federated setting to evaluate the performance. 
In Table \ref{Tab:CIFAR-LT}, we give tha results on both the IID and non-IID data settings built upon the  global long-tailed dataset CIFAR-10-LT with different imbalance factors 10, 50 and 100. For non-IID data partition, we use Dirichlet distribution-based sampling method with different concentration parameter $\alpha$ to control the degree of data heterogeneity.
To better demonstrate the impacts of the long-tailed data distribution, we also include a group of experiment results on the (balanced) CIFAR-10 for reference. 
In Table \ref{CIFAR}, results on CIFAR-10 are provided, where we consider sample different long-tailed local data distributions (i.e., different head-tail distribution) with the same imbalance factor IF$_\text{L}$. See Figure \ref{fig:heatmap}(C) for an overview.

For the results in Tables \ref{Tab:CIFAR-LT} and  \ref{CIFAR}, best test accuracies of all algorithms present a descending sort pattern from the left to right,  as the degree of the long-tail and heterogeneity is increasing. Interestingly, the federated optimization methods FedProx outperforms FedAvg in the non-long-tailed setting, while it tends to underperform with global long-tailed data in some settings. As a specific method to tackle long-tailed data, CReFF can achieve best results among all four algorithms in most of settings, but it has lower accuracy performances than FedProx with more heterogeneous data distribution.
With regard to the PFL methods, our preliminary results illustrate that personalization method outperforms in most of the long-tailed  data settings, especially in settings of Table \ref{CIFAR} (i.e., diverse local long-tailed distributions in Type 2). 

The numerical results indicate that, PFL methods have the potential to enhance the performance without any specialized long-tailed learning techniques. 
More importantly, the preliminary results also demonstrate the feasibility and possibility to re-purpose the federated optimization and PFL methods with centralized long-tailed learning approaches in federated scenarios.

\section{Future Trends and Research Opportunities}
\label{Sec:Future}
Based on the above experimental results and discussions of the federated long-tailed learning, we envision the following directions and opportunities towards the robust and communication-efficient federated long-tailed learning algorithms, architectures and analysis.
\begin{itemize}
    \item \textbf{Incorporate PFL ideas for better federated long-tail learning.} As a promising technique, PFL could possibly boost the training performance of federated long-tailed learning with centralized long-tailed learning methods. How to balance the global shared knowledge with local perosnalized knowledge could be incorporated into the design of the representation learning and classification architectures in federated long-tailed learning.
    Moreover, it would be promising to explore the incorporation of the model-based and data-based PFL approaches \cite{tan2022towardspfl} with the long-tailed learning.
    \item \textbf{Hierarchical FL architectures. } In the presence of diverse data distributions, we may consider to group clients with similar long-tail distributional statistics into clusters to jointly learn cluster-level personalized models or conduct cluster-level MTL \cite{sattler2020clustermtl}. However, the design of a privacy-preserving clustering method remains to be further investigated.
    \item \textbf{Re-purpose of existing federated optimization methods.} Local long-tailed data distribution could be regarded as an extremely imbalanced case of  data heterogeneity. Hence, how to re-purpose the federated optimization algorithm in the presence of the long-tailed data could be further explored.
    It would be another open question to develop a heterogeneity-agnostic federated optimization framework. 
    Moreover, MTL-based long-tailed learning could also be a potential approach to address the heterogeneous long-tailed distributions in FL.
    \item \textbf{Design better data partition/sampling schemes or more representative datasets.} In addition to the several real-world long-tailed datasets, most of the current work use the long-tailed version of the popular image datasets. Although this method could use the pre-determined imbalance factor IF$_\text{G}$ to control the imbalance, it would also discard a large amount of samples when following the widely-used exponential and Pareto sampling methods. Therefore, the degradation of the performance could also be partially attributed to the small dataset size, especially for scenarios  with   a larger imbalance factor in federated settings. How to mitigate such negative impacts should be further investigated. Meanwhile, future research could also leverage on real-world scenarios, such as medical images or autonomous cars, to provide more representative and convincing federated long-tailed learning dataset. 
\end{itemize}

\section{Concluding Remarks}

In this paper, we introduce the federated long-tailed learning task, a general setting motivated by real-world applications but rarely studied in previous research. We characterize three types of F-LT learning settings with diverse local and global long-tailed data distributions. The benchmark results with multiple federated learning architectures suggest that substantial future work is needed for better F-LT. In addition, we highlight the potential techniques and possible trajectories of research towards federated long-tailed learning with real-world data.  

\appendix


\bibliographystyle{named}
\bibliography{ijcai22}

\end{document}